\documentclass[10pt,twocolumn]{ICCAS}
 
\usepackage{diagbox}
\usepackage{multirow}
\usepackage{cite}
\usepackage{float}

\begin{document}

\title{Beer-Lambert Guided Representation Learning for Unsupervised Anomaly Detection in Sub-THz Food Inspection Images}

\author{Gyutae Hwang and Sang Jun Lee${}^{*}$ }

\affils{Division of Electronics and Information Engineering, Jeonbuk National University, \\
567 Baekje-daero, Deokjin-gu, Jeonju, 54896, Jeonbuk-do, Republic of Korea \\ 
(gyutae741@jbnu.ac.kr) (sj.lee@jbnu.ac.kr) {\small${}^{*}$ Corresponding author}}


\abstract{
Food manufacturing requires reliable inspection systems to detect foreign material contamination and maintain product safety.
Sub-THz transmission imaging provides material-dependent attenuation characteristics that are useful for detecting low-density contaminants in food products.
However, existing unsupervised anomaly detection methods mainly rely on RGB-pretrained visual representations, which may not adequately capture the transmission behavior of Sub-THz images.
This paper proposes a Beer-Lambert guided representation learning framework for unsupervised anomaly detection in Sub-THz food inspection images.
The proposed method introduces an attenuation decomposition module as an auxiliary regularization module that constrains student representations through attenuation reconstruction during training.
In addition to the conventional one-class setting, we introduce a Leave-One-Food-Out protocol to evaluate generalization capability under unseen food categories.
Experimental results on the Inline-Food-Inspection-THz dataset show that the proposed method improves overall anomaly detection performance over the baseline method.
}


\maketitle


\section{Introduction}
Food manufacturing processes require reliable quality control systems to ensure product safety and maintain consumer trust. 
Among various quality assurance tasks, the detection of foreign material contamination remains one of the most critical challenges in industrial production environments. 
Contaminants can be introduced during processing, packaging, or transportation, resulting in significant economic losses and potential safety risks. 
To address this issue, non-destructive inspection technologies such as hyperspectral imaging (HSI) and X-ray imaging have been widely adopted for automated food inspection systems~\cite{fengyz2012applicationofhyperspectralimaginginfood,haff2008x}. 
However, conventional imaging modalities often exhibit limited sensitivity to low-density contaminants embedded within food products. 
Recently, Sub-THz imaging has emerged as a promising inspection modality by providing access to internal transmission characteristics of food products, thereby improving the detection ability of low-density contaminants~\cite{papadopoulos2026detection}.  

\begin{figure}[t]
\begin{center}
\includegraphics[width=\columnwidth]{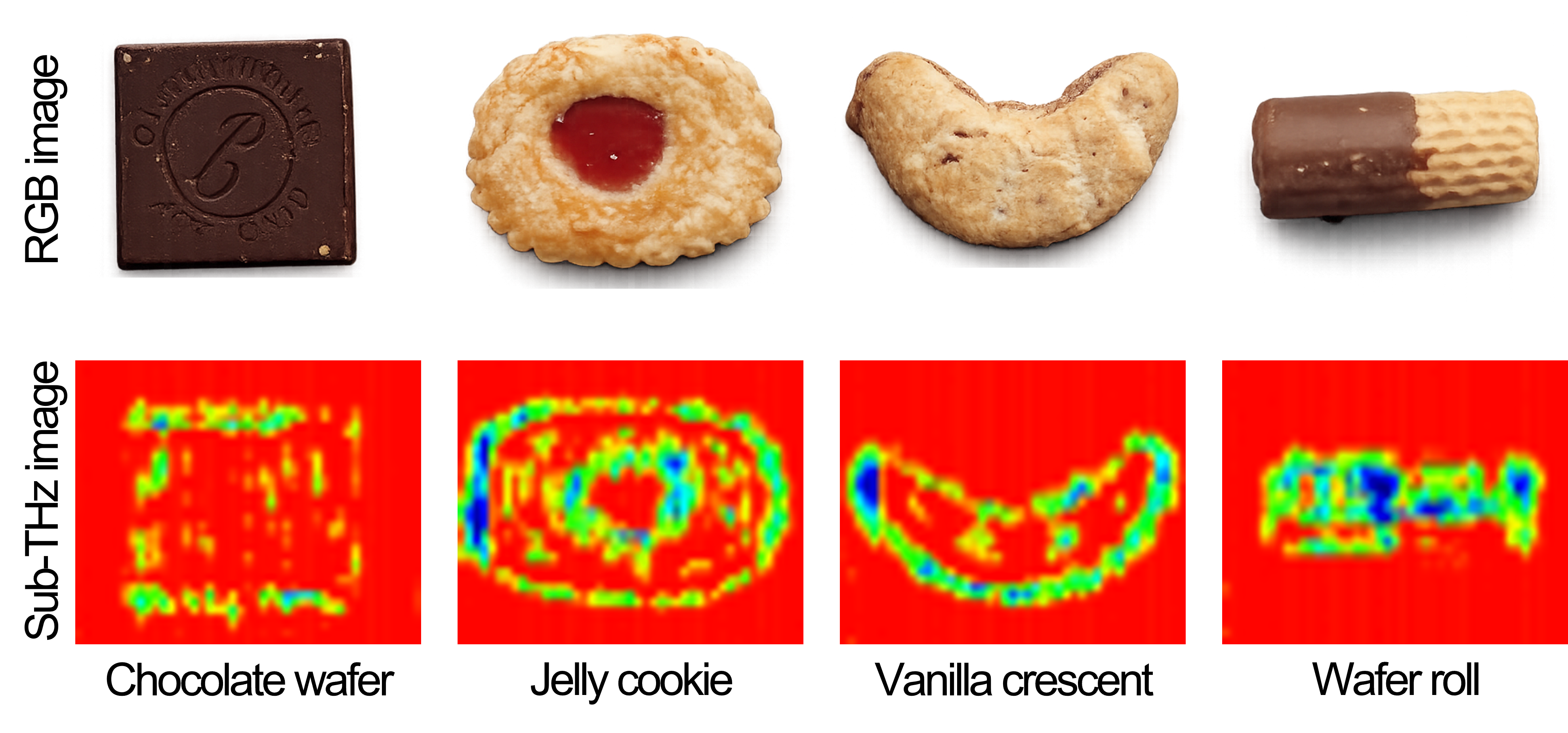}
\caption{\label{fig.1} Representative samples from the Inline-Food-Inspection-THz dataset. The top and bottom rows show RGB images and the corresponding Sub-THz transmission images, respectively.}
\end{center}
\end{figure}

Unlike RGB images, Sub-THz transmission images primarily reflect material-dependent attenuation characteristics rather than surface appearance.
Food products are composed of heterogeneous materials with varying absorption and transmission properties. 
As shown in Fig. \ref{fig.1}, the intensity observed in Sub-THz transmission food images is determined by the interaction between electromagnetic waves and the constituent materials.
According to the Beer-Lambert law, the transmitted intensity decreases exponentially as electromagnetic waves propagate through an absorbing material, where the attenuation is determined by the material properties and propagation distance~\cite{beer1852}. 
Consequently, attenuation maps derived from transmission measurements provide physically meaningful information regarding the internal composition and structural characteristics.
These discriminative representations can be used to detect anomalous contaminants in Sub-THz food inspection images.
Since anomalous data samples are rarely observed in real-world food manufacturing environments, unsupervised anomaly detection (UAD) provides a practical framework for learning normal patterns from Sub-THz food inspection images.

Recent advances in deep representation learning and large-scale visual pretraining~\cite{caron2021emerging} have significantly improved the performance of UAD in industrial inspection tasks.
Previous UAD methods~\cite{defard2021padim, roth2022towards, deng2022anomaly, batzner2024efficientad, rolih2024supersimplenet, wei2025uninet} learn discriminative representations of normal samples and detect anomalies through feature discrepancies or reconstruction errors. 
These methods have demonstrated remarkable performance on industrial UAD benchmarks such as MVTec AD~\cite{bergmann2019mvtec} and VisA~\cite{zou2022spot}. 
In particular, state-of-the-art approaches leverage feature extractors pretrained on large-scale RGB image datasets to obtain robust visual representations. 
However, such RGB-aware pretrained parameters are optimized for natural image statistics and may not adequately capture the transmission characteristics of Sub-THz images. 
Consequently, existing UAD methods do not explicitly exploit attenuation information derived from electromagnetic wave propagation, which may limit their robustness under variations in food types, thickness, and transmission properties.

To address these limitations, this paper proposes a Beer-Lambert guided representation learning framework for UAD in Sub-THz food inspection images. 
Although attenuation information provides physically meaningful cues for anomaly detection, the publicly available Inline-Food-Inspection-THz dataset only provides preprocessed grayscale images. 
Inspired by the Beer-Lambert law, we model attenuation as the interaction between absorption characteristics and propagation distance.
The proposed method learns absorption and thickness maps and factorizes them to reconstruct attenuation maps from Sub-THz transmission images.
By preserving the attenuation characteristics of normal food products during training, the encoder learns representations for anomaly detection that better reflect Sub-THz transmission and attenuation behavior.
Furthermore, we introduce a Leave-One-Food-Out (LOFO) protocol in addition to the conventional one-class setting to evaluate generalization capability across different food products. 
Experiments on the Inline-Food-Inspection-THz dataset establish benchmark results for existing UAD methods and demonstrate the effectiveness of the proposed method.

\section{Materials}

\subsection{Inline-Food-Inspection-THz Dataset}

The Inline-Food-Inspection-THz dataset consists of Sub-THz transmission images collected for food contamination inspection under diverse foreign material conditions. 
The dataset includes four representative food products: chocolate wafer, jelly cookie, vanilla crescent, and wafer roll. 
For each food product, both reference and contaminated samples were acquired under controlled measurement settings. 
The contamination scenarios include six foreign materials: metal, glass, polyethylene (PE), polypropylene (PP), polystyrene (PS), and polyvinyl chloride (PVC), with sizes of 5 mm and 7 mm.
To incorporate variations in transmission characteristics, measurements were performed under multiple paper thickness conditions ranging from 0p to 20p. 
The total dataset contains 288 Sub-THz images, including both normal and contaminated samples across all food categories and acquisition conditions. 
These diverse acquisition conditions provide a benchmark for evaluating anomaly detection methods in Sub-THz food inspection.

\subsection{Evaluation protocols}

To evaluate both anomaly detection performance and generalization capability, we employed two evaluation protocols, one-class anomaly detection and LOFO testing.
For one-class anomaly detection, experiments were conducted independently for each food category.
Normal samples acquired under the paper thickness 0p and 20p were used for training, resulting in 24 training images per category. 
The test set consisted of 12 normal samples from the 10p condition and 36 contaminated samples collected under all paper thickness conditions. 
This protocol evaluates the ability of a model to detect contaminated samples within a single food category under unseen measurement conditions.

To evaluate generalization capability, we additionally introduce a LOFO protocol. 
In each fold, a food category was held out for testing, while the remaining food categories were used for model training. 
This configuration yields 108 training images and 64 test images for each fold. 
Unlike the conventional one-class setting, the LOFO protocol trains an UAD model on multiple food products and evaluates whether the learned Sub-THz representations can generalize to an unseen product without additional adaptation.

\begin{figure*}[t]
\begin{center}
\includegraphics[width=\linewidth]{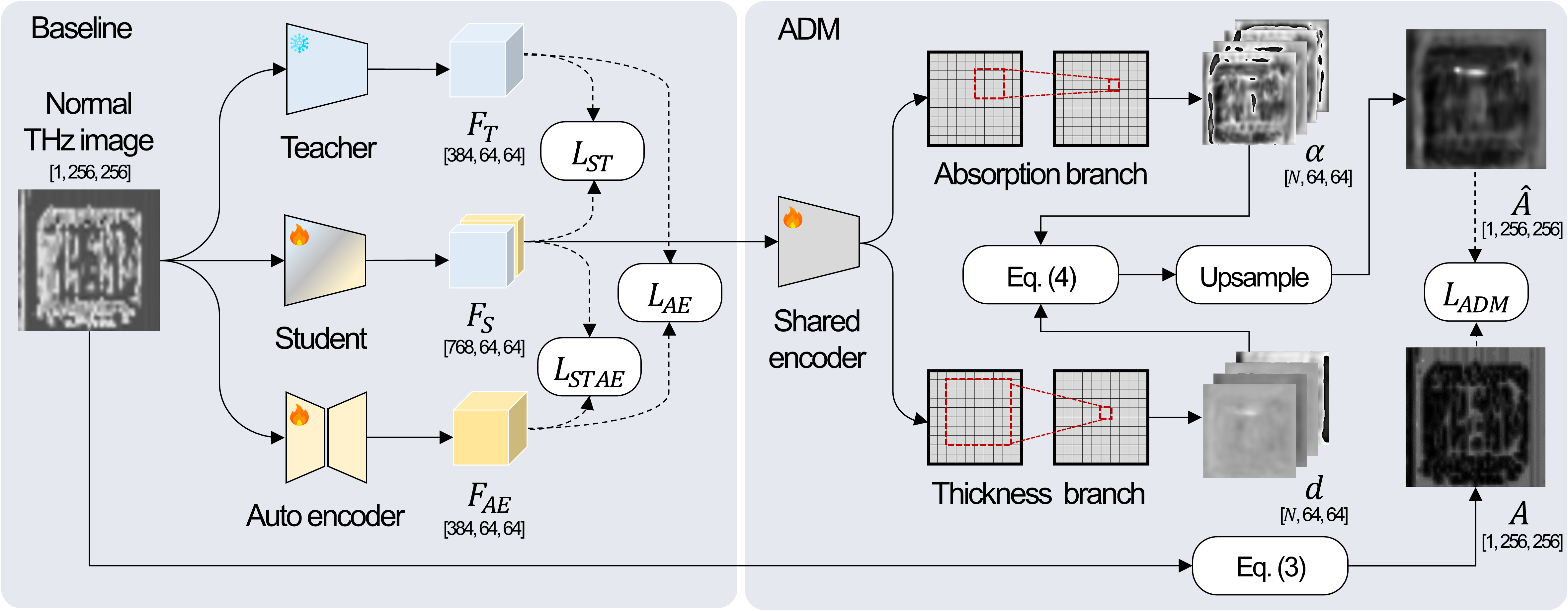}
\caption{\label{fig.2} Training pipeline of the proposed method. Solid lines represent feature propagation, whereas dotted lines denote the training losses.}
\end{center}
\end{figure*}

\section{Methodology}
\subsection{Beer-Lambert guided attenuation modeling}
\label{beerlambert}
Sub-THz transmission images are formed through the interaction between electromagnetic waves and the internal composition of food products.
Unlike RGB images that primarily capture surface appearance, the intensity observed in Sub-THz images is governed by the absorption and transmission characteristics of the constituent materials.
Since food products consist of multiple heterogeneous material layers, the observed attenuation reflects the cumulative absorption effect along the propagation path.
According to the Beer-Lambert law, the transmitted intensity $I$ is related to the incident intensity $I_0$ by

\begin{equation}
I = I_0 \exp(-A),
\label{eq:beer_lambert}
\end{equation}

\noindent where $A$ denotes the attenuation accumulated along the propagation path. For a heterogeneous structure composed of multiple material layers, the attenuation can be expressed as

\begin{equation}
A = \sum_{i=1}^{N} \alpha_i d_i,
\label{eq:attenuation_sum}
\end{equation}

\noindent where $\alpha_i$ and $d_i$ denote the absorption coefficient and propagation distance of the $i$-th material layer, respectively. 
This formulation indicates that the observed attenuation is jointly determined by the material-dependent absorption characteristics and the propagation distance.
By applying a logarithmic transformation to the normalized transmission intensity, an attenuation map can be obtained as

\begin{equation}
A = -\log \left( \frac{I}{I_0} \right).
\label{eq:attenuation_map}
\end{equation}

\noindent The resulting attenuation map provides physically meaningful information regarding the internal composition and structural characteristics of food products.
Therefore, attenuation maps offer a more suitable representation for Sub-THz image than appearance-based image features alone.
Motivated by this observation, we incorporate attenuation modeling into the representation learning process. 
Rather than explicitly estimating the attenuation contribution of each material layer, the proposed framework learns absorption and thickness representations whose interaction approximates the attenuation characteristics observed in Sub-THz transmission images.

\subsection{Baseline architecture}
Recent UAD methods heavily rely on representations obtained from large-scale pretrained models to characterize defect-free feature distributions in discriminative feature spaces. 
EfficientAD is a lightweight UAD framework that consists of a frozen teacher network, a trainable student network, and an autoencoder. 
The architecture of EfficientAD is illustrated on the left side of Fig.~\ref{fig.2}.
Given a normal Sub-THz image, the teacher extracts reference features $F_T$, while the student learns to mimic these representations through feature distillation. 
The student-teacher loss $L_{ST}$ minimizes the discrepancy between $F_T$ and $F_S$, encouraging the student to model the distribution of normal samples. 
In parallel, the autoencoder is trained to reconstruct teacher representations with the autoencoder loss $L_{AE}$, while the student-autoencoder loss $L_{STAE}$ further constrains feature consistency among the networks.
All training objectives in EfficientAD are formulated using pixel-wise $L_2$ distances between feature representations. 
This distillation strategy detects anomalies as feature discrepancies between the fixed teacher and the student adapted to defect-free samples.
Despite its effectiveness, EfficientAD learns representations primarily from pretrained RGB feature priors without explicitly incorporating the attenuation characteristics that govern Sub-THz image formation. 
To address this limitation, the proposed method introduces a Beer-Lambert guided attenuation decomposition module into the student representation learning process.

\subsection{Attenuation decomposition module}

The attenuation decomposition module (ADM) is designed to constrain representation learning to preserve attenuation characteristics derived from the Beer-Lambert law.
The proposed module estimates factors corresponding to absorption and thickness from the student representation to reconstruct the attenuation map.
Since the Inline-Food-Inspection-THz dataset does not provide absorption coefficients, material thicknesses, or attenuation annotations, these quantities cannot be directly supervised during training.
Instead, the ADM encourages the network to infer absorption and thickness representations that explain the heterogeneous attenuation structure observed in normal Sub-THz images.
Through attenuation reconstruction, the encoder learns representations that explain the attenuation structure of normal samples.
As a result, attenuation patterns introduced by foreign materials are difficult to represent from the learned decomposition features, leading to larger reconstruction errors and feature discrepancies.

As shown on the right side of Fig.~\ref{fig.2}, the student feature $F_S$ is first processed by a shared encoder.
The shared encoder consists of convolutional layers that generate a compact representation shared by the absorption branch and the thickness branch.
The absorption branch employs two $3 \times 3$ convolutions to estimate $N$ absorption maps $\alpha$.
These maps capture local high-frequency attenuation variations that reflect material-dependent absorption properties.
The thickness branch employs $7 \times 7$ convolutions followed by average pooling to estimate $N$ thickness maps $d$.
The larger receptive field suppresses high-frequency fluctuations and captures low-frequency structural information related to propagation distance and food thickness.
The attenuation map $\hat{A}$ is reconstructed from the absorption and thickness maps as follows:

\begin{equation}
\hat{A}
=
\mathrm{Upsample}
\left(
\sum_{n=1}^{N}
\alpha_n \odot d_n
\right),
\label{eq:factorization}
\end{equation}

\noindent where $\odot$ denotes element-wise multiplication.
The reconstructed attenuation feature is subsequently upsampled to obtain $\hat{A}$.
The target attenuation map $A$ is computed using Eq.~(\ref{eq:attenuation_map}).
The resulting attenuation representation reflects the cumulative absorption effects of heterogeneous materials and is consistent with the Beer-Lambert formulation described in Section 3.1.
The discrepancy between the reconstructed attenuation map $\hat{A}$ and the target attenuation map $A$ is minimized by the ADM loss

\begin{equation}
L_{ADM}=\left|\hat{A}-A\right|_1,
\label{eq:ladm_loss}
\end{equation}

\noindent which acts as a regularization term that guides the student features toward meaningful transmission patterns.
The proposed method is trained jointly with the original EfficientAD objectives.

\begin{equation}
L=L_{ST}+L_{AE}+L_{STAE}+\lambda L_{ADM},
\label{eq:total_loss}
\end{equation}

\noindent where $\lambda$ indicates weighting coefficients.

\subsection{Anomaly score}

During inference, the proposed method follows the anomaly scoring strategy of EfficientAD.
The student feature consists of two feature groups corresponding to the student-teacher and student-autoencoder objectives.
The teacher-student and student-autoencoder anomaly maps are computed as

\begin{equation}
M_{ST}=\left|F_T-F_{S,T}\right|_2^2,
M_{AE}=\left|F_{AE}-F_{S,AE}\right|_2^2,
\label{eq:m}
\end{equation}

\noindent where $F_{S,T}$ and $F_{S,AE}$ denote the student features used for teacher distillation and autoencoder alignment, respectively.
The final anomaly map $M$ and image-level anomaly score $S$ are obtained as

\begin{equation}
M = M_{ST} + M_{AE},
\qquad S = \max(M).
\label{eq:anomaly_score}
\end{equation}

\noindent The ADM is used only for representation learning and removed during inference, preserving the computational efficiency of the original EfficientAD framework.

\section{Experimental results}
\subsection{Experimental settings and evaluation metrics}

All experiments were conducted on a hardware system equipped with an Intel Core i9-10940X CPU, 64 GB DDR4 RAM, and dual NVIDIA GeForce RTX 3090 Ti GPUs. 
The software environment consisted of Ubuntu 20.04, Python 3.10, PyTorch 2.4.1, and Anomalib 2.4.2~\cite{akcay2022anomalib}. 
The proposed model was trained using the Adam optimizer with a batch size of 1, a learning rate of $1\times10^{-4}$, and a weight decay of $1\times10^{-5}$. 
The ADM was configured with four attenuation components and an attenuation reconstruction loss weight of $\lambda=0.1$. 
All models were trained for a fixed 100 epochs without early stopping. 
Image-level AUROC and F1-score were used to evaluate anomaly detection performance.

\subsection{One-class anomaly detection}

Table~\ref{tab:oneclass} presents the anomaly detection performance under the conventional one-class setting. 
The results show that distillation-based methods generally outperform memory-bank-based approaches on the Inline-Food-Inspection-THz dataset. 
PatchCore and PaDiM rely on patch-level feature matching and are sensitive to the limited number of normal training samples available in the dataset. 
In contrast, EfficientAD and proposed method learn compact representations through teacher-guided feature learning and achieve higher overall performance.
The proposed method further extends EfficientAD by introducing attenuation-aware supervision derived from the Beer-Lambert law. 
Compared with EfficientAD, the proposed method improves the total AUROC from 0.983 to 0.986 and the total F1-score from 0.959 to 0.968, corresponding to gains of 0.3 and 0.9 percentage points, respectively. 
These results indicate that attenuation-aware supervision improves feature discrimination by preserving physically meaningful attenuation characteristics in addition to feature consistency.

\subsection{Leave-one-food-out evaluation}

Table~\ref{tab:lofo} presents the anomaly detection results under the proposed LOFO protocol. 
Compared with the conventional one-class setting, most existing methods exhibited substantial performance degradation under the LOFO setting. 
Memory-bank-based methods showed the largest performance degradation due to the stored normal representations were strongly coupled to food-specific appearance characteristics. 
Although distillation-based methods exhibited improved robustness, their representations were still learned primarily from RGB-pretrained features optimized for natural image statistics. 
The proposed method introduces attenuation-aware supervision and encourages the network to learn intrinsic transmission characteristics of Sub-THz images rather than food-specific appearance patterns. 
Compared with EfficientAD, the proposed method improves the total AUROC from 0.865 to 0.884 and the total F1-score from 0.840 to 0.879, corresponding to gains of 1.9 and 3.9 percentage points, respectively. 
These results indicate that Beer-Lambert guided representation learning improves generalization capability by capturing attenuation characteristics that are less dependent on food category.

\begin{table*}[t]
\centering
\caption{Performance comparison under the one-class setting. The best and second-best results are shown in bold and underline, respectively.}
\label{tab:oneclass}
\setlength{\tabcolsep}{3pt}
\renewcommand{\arraystretch}{1.1}
\resizebox{0.94\textwidth}{!}{
\begin{tabular}{lc|cc|cc|cc|cc|cc}
\hline
\multirow{2}{*}{Method} & \multirow{2}{*}{Year}
& \multicolumn{2}{c|}{Chocolatewafer}
& \multicolumn{2}{c|}{Jellycookie}
& \multicolumn{2}{c|}{Vanillacrescent}
& \multicolumn{2}{c|}{Waferroll}
& \multicolumn{2}{c}{Total} \\
\cline{3-12}
& & AUROC & F1 & AUROC & F1 & AUROC & F1 & AUROC & F1 & AUROC & F1 \\
\hline
PaDiM~\cite{defard2021padim} & 2021
& 0.949 & \textbf{0.972}
& 0.851 & 0.932
& 0.924 & 0.959
& \underline{0.998} & \underline{0.972}
& 0.930 & 0.959 \\

PatchCore~\cite{roth2022towards} & 2022
& \textbf{0.993} & \underline{0.959}
& 0.961 & 0.959
& 0.956 & 0.959
& \textbf{1.000} & \textbf{0.986}
& 0.977 & \underline{0.966} \\

Reverse Distillation~\cite{deng2022anomaly} & 2022
& \textbf{0.993} & \underline{0.959}
& 0.991 & \underline{0.972}
& 0.679 & 0.843
& \textbf{1.000} & \textbf{0.986}
& 0.916 & 0.940 \\

EfficientAD~\cite{batzner2024efficientad} & 2024
& 0.940 & 0.907
& \underline{0.993} & \underline{0.959}
& \textbf{1.000} & \textbf{0.986}
& \textbf{1.000} & \textbf{0.986}
& \underline{0.983} & 0.959 \\

SupersimpleNet~\cite{rolih2024supersimplenet} & 2024
& 0.641 & 0.843
& 0.047 & 0.843
& 0.567 & 0.843
& 0.050 & 0.843
& 0.326 & 0.843 \\

UniNet~\cite{wei2025uninet} & 2025
& 0.898 & 0.907
& 0.694 & 0.864
& 0.697 & 0.843
& 0.687 & 0.843
& 0.744 & 0.864 \\
\hline
\textbf{Ours} & 
& \underline{0.951} & 0.930
& \textbf{1.000} & \textbf{0.986}
& \underline{0.993} & \underline{0.972}
& \textbf{1.000} & \textbf{0.986}
& \textbf{0.986} & \textbf{0.968} \\
\hline
\end{tabular}
}
\end{table*}

\begin{table*}[t]
\vspace{2mm}
\centering
\caption{Performance comparison under the LOFO setting. The best and second-best results are shown in bold and underline, respectively.}
\label{tab:lofo}
\setlength{\tabcolsep}{3pt}
\renewcommand{\arraystretch}{1.1}
\resizebox{0.94\textwidth}{!}{
\begin{tabular}{lc|cc|cc|cc|cc|cc}
\hline
\multirow{2}{*}{Method} & \multirow{2}{*}{Year}
& \multicolumn{2}{c|}{Chocolatewafer}
& \multicolumn{2}{c|}{Jellycookie}
& \multicolumn{2}{c|}{Vanillacrescent}
& \multicolumn{2}{c|}{Waferroll}
& \multicolumn{2}{c}{Total}
\\
\cline{3-12}
&
& AUROC & F1
& AUROC & F1
& AUROC & F1
& AUROC & F1
& AUROC & F1 \\
\hline

PaDiM~\cite{defard2021padim} & 2021
& 0.546 & 0.695
& 0.725 & 0.757
& 0.785 & 0.735
& 0.431 & 0.660
& 0.622 & 0.712
 \\

PatchCore~\cite{roth2022towards} & 2022
& 0.712 & 0.707
& 0.812 & 0.759
& 0.721 & 0.703
& \underline{0.725} & 0.732
& 0.743 & 0.725
 \\

Reverse Distillation~\cite{deng2022anomaly} & 2022
& 0.558 & 0.742
& 0.829 & 0.753
& 0.767 & 0.740
& 0.717 & \underline{0.745}
& 0.718 & 0.745
 \\

EfficientAD~\cite{batzner2024efficientad} & 2024
& \underline{0.988} & \underline{0.932}
& \textbf{0.975} & \textbf{0.925}
& \underline{0.910} & \underline{0.849}
& 0.586 & 0.654
& \underline{0.865} & \underline{0.840}
 \\

SupersimpleNet~\cite{rolih2024supersimplenet} & 2024
& 0.429 & 0.660
& 0.606 & 0.681
& 0.837 & 0.789
& \textbf{0.807} & 0.729
& 0.670 & 0.715
 \\

UniNet~\cite{wei2025uninet} & 2025
& 0.525 & 0.660
& 0.441 & 0.654
& 0.314 & 0.654
& 0.409 & 0.654
& 0.422 & 0.655
 \\
\hline

\textbf{Ours} & 
& \textbf{0.996} & \textbf{0.958}
& \underline{0.935} & \underline{0.917}
& \textbf{0.938} & \textbf{0.883}
& 0.665 & \textbf{0.759}
& \textbf{0.884} & \textbf{0.879}
\\
\hline

\end{tabular}
}
\end{table*}

\subsection{Qualitative results}
\begin{figure}[H]
\begin{center}
\includegraphics[width=\columnwidth]{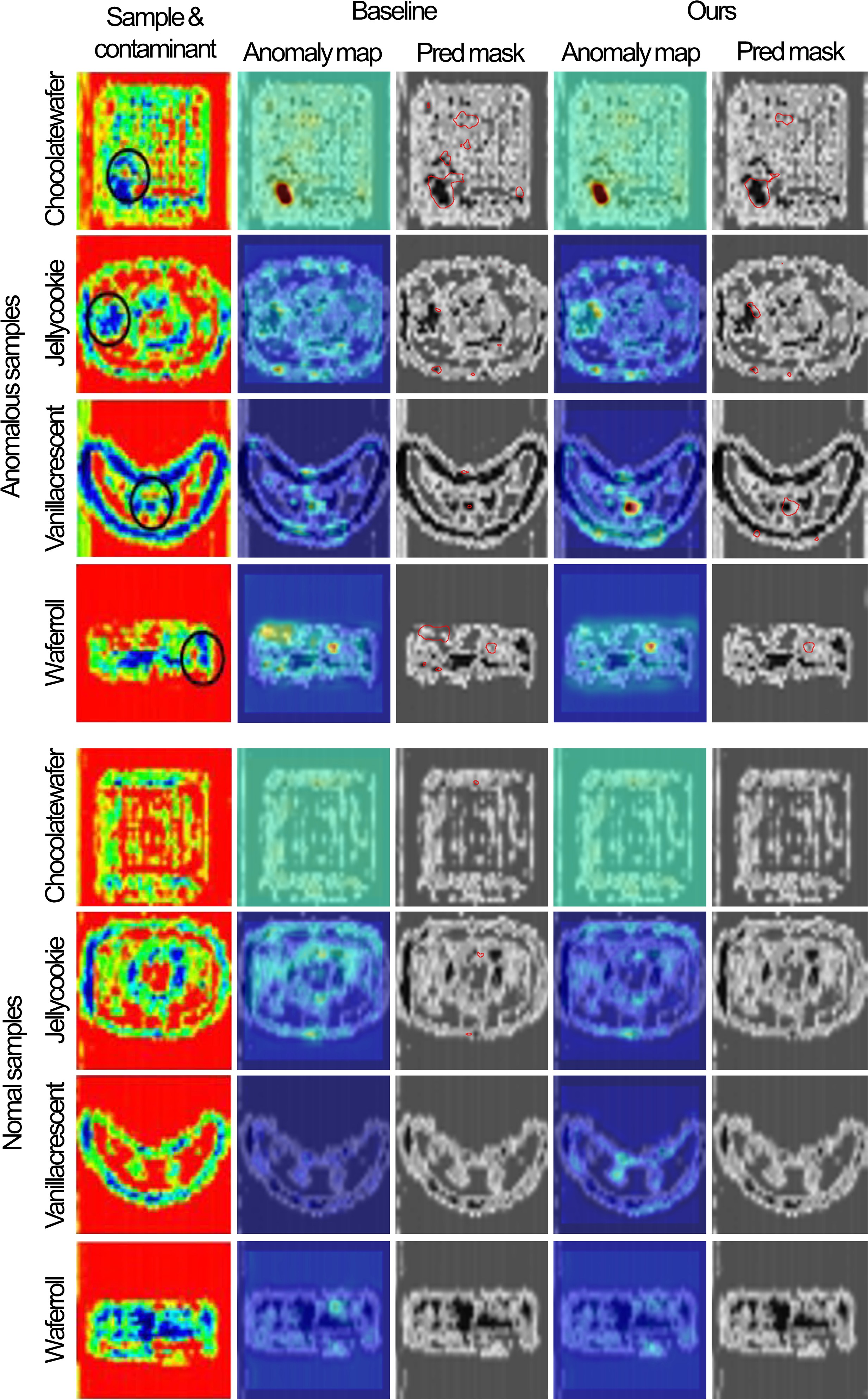}
\caption{\label{fig:qualitative}
Qualitative comparison. Contaminated regions are highlighted by black circles.}
\end{center}
\end{figure}

We visualize the anomaly maps and prediction masks generated by the baseline and the proposed method, as shown in Fig.~\ref{fig:qualitative}. 
Although the Inline-Food-Inspection-THz dataset does not provide pixel-level anomaly annotations, the original dataset paper reports the insertion locations of foreign materials. 
Therefore, qualitative evaluation of anomaly localization is possible by comparing the anomaly responses with the reported contamination locations. 
As shown in Fig.~\ref{fig:qualitative}, the baseline often produces diffuse anomaly responses over broad regions of the food sample. 
In contrast, the proposed method generates more localized anomaly responses around the contaminated regions highlighted by black circles. 
For normal samples, the proposed method suppresses unnecessary activations and produces cleaner anomaly maps. 
These results suggest that attenuation-aware representation learning encourages the network to focus on transmission-related abnormal patterns rather than food-specific appearance characteristics.

\section{Conclusion}
This paper presented a Beer-Lambert guided representation learning framework for unsupervised anomaly detection in Sub-THz food inspection images. 
The proposed method introduced a ADM that learns absorption and thickness representations to reconstruct attenuation maps from transmission images. 
By incorporating attenuation-aware supervision into EfficientAD, the proposed framework encourages the network to preserve physically meaningful transmission characteristics during training. 
Experimental results on the Inline-Food-Inspection-THz dataset demonstrated consistent improvements over the baseline method under both one-class and LOFO settings. 
In particular, the proposed method achieved superior cross-food generalization performance on unseen food categories. 
Qualitative results further showed more localized anomaly responses around contaminated regions while suppressing unnecessary activations in normal samples. 
These results demonstrate the effectiveness of Beer-Lambert guided representation learning for robust anomaly detection in Sub-THz food inspection.

\section*{ACKNOWLEDGEMENT}
This work was supported by the Institute of Information \& Communications Technology Planning \& Evaluation(IITP)-Innovative Human Resource Development for Local Intellectualization program grant funded by the Korea government(MSIT)(IITP-2026-RS-2024-00439292)

%

%

\end{document}